\documentclass[a4paper]{article}
\usepackage{INTERSPEECH2019}
\usepackage{amsmath,graphicx,hyperref,typeface,float,enumitem}
\usepackage{multirow}

\title{Exploiting Syntactic Features in a Parsed Tree \\to Improve End-to-End TTS}
\name{Haohan Guo$^*$, Frank K. Soong$^\dag$, Lei He$^\dag$, Lei Xie$^*$\thanks{$^*$Work performed as an intern at Microsoft.}}
\address{$^*$School of Computer Science, Northwestern Polytechnical University, Xi’an, China \\
  $^\dag$Microsoft AI \& Research, Beijing, China}
\email{\{hhguo,lxie\}@nwpu-aslp.org, \{frankkps, helei\}@microsoft.com}

\begin{document}
%
\maketitle
\begin{abstract}
The end-to-end TTS, which can predict speech directly from a given sequence of graphemes or phonemes, has shown improved performance over the conventional TTS. However, its predicting capability is still limited by the acoustic/phonetic coverage of the training data, usually constrained by the training set size. To further improve the TTS quality in pronunciation, prosody and perceived naturalness, we propose to exploit the information embedded in a syntactically parsed tree where the inter-phrase/word information of a sentence is organized in a multilevel tree structure. Specifically, two key features: phrase structure and relations between adjacent words are investigated. Experimental results in subjective listening, measured on three test sets, show that the proposed approach is effective to improve the pronunciation clarity, prosody and naturalness of the synthesized speech of the baseline system.
\end{abstract}


\section{Introduction}
\label{sec:intro}

Evaluation of text-to-speech (TTS) system focuses on measuring several factors in intelligibility, naturalness, prosody and speaker similarity. Conventional speech parameter-based TTS system has achieved high intelligibility, e.g., GMM-HMM-based \cite{tokuda2000speech} and NN-based \cite{ze2013statistical, zen2015acoustic, fan2014tts} statistical speech synthesis. Recently, the speech quality has also been greatly improved in the WaveNet \cite{van2016wavenet, tamamori2017speaker} or WaveRNN \cite{kalchbrenner2018efficient, valin2018lpcnet} based neural vocoder, which can produce high quality speech by predicting speech samples from the generated acoustic features. However, its predicting capability in pronunciation, prosody and naturalness is still limited by its acoustic/phonetic coverage and the amount of data available for training. 

In conventional English TTS, ToBI labels are often used to transcribe the prosody change, including: stress, emphasis, breaks, etc.  Speech prosody of the training data can be annotated in ToBI \cite{silverman1992tobi} manually or automatically \cite{rosenberg2010autobi} first to train a ToBI prosody model for predicting ToBI labels from the given text. Annotation is based on text and audio, but in prediction only text is available. A high quality ToBI prosody model, which needs to take long-term context into account to predict the prosody, can be difficult to train with limited, text-speech paired data. In addition, the models which need to predict the duration, break, voicing and F0 contours can make the training even more challenging. Prediction errors of the prosody model can be accumulated, then degrade the prediction of the spectral parameters for TTS and lead to unexpected glitches in synthesized speech. Besides, prosody information cannot be fully characterized by the ToBI label sequence.

Recently, end-to-end TTS training was proposed, e. g. char2wav \cite{sotelo2017char2wav}, Tacotron \cite{wang2017tacotron} and Tacotron2 \cite{shen2018natural} to predict speech parameters directly from graphemes or phonemes in a unified way and there is no need to manually annotate speech data to train the model. It can learn various acoustic patterns via a flexible mapping between the linguistic space to acoustic space by minimizing the prediction error in the iterative training loop. All modules in the end-to-end model are trained jointly, so the accumulated errors caused by separated training module can thus be avoided. Experimental results show that an end-to-end model performs better than the conventional, statistical TTS. However, problems can still occur, e.g. wrong stress patterns, unreasonable breaks, mispronunciations, especially for long and complex test sentences outside the domain covered by the training data. The resultant poor generalization can seriously degrade the corresponding TTS performance.

End-to-end model is a sequence-to-sequence model which is highly dependent on sequential information. But the size of training sentences is not enough to cover the target (text) domain, including different length and context. To improve the generalization capability of the model, we need to improve the coverage of our data on text domain as much as possible, and the best way is to improve the generalization of data. The sequences which are only composed of graphemes or phonemes have low generalization because every sequence refers to the specific case, which can't represent some cases sharing the common features. It leads to the problem that many cases are not well covered in the training set. So we can use higher-level and more abstract features to describe the sequence to improve the coverage and generalization of input data. Semantic information and syntactic information are what we need.

\begin{figure}[htp]
  \centering
  \includegraphics[width=8cm]{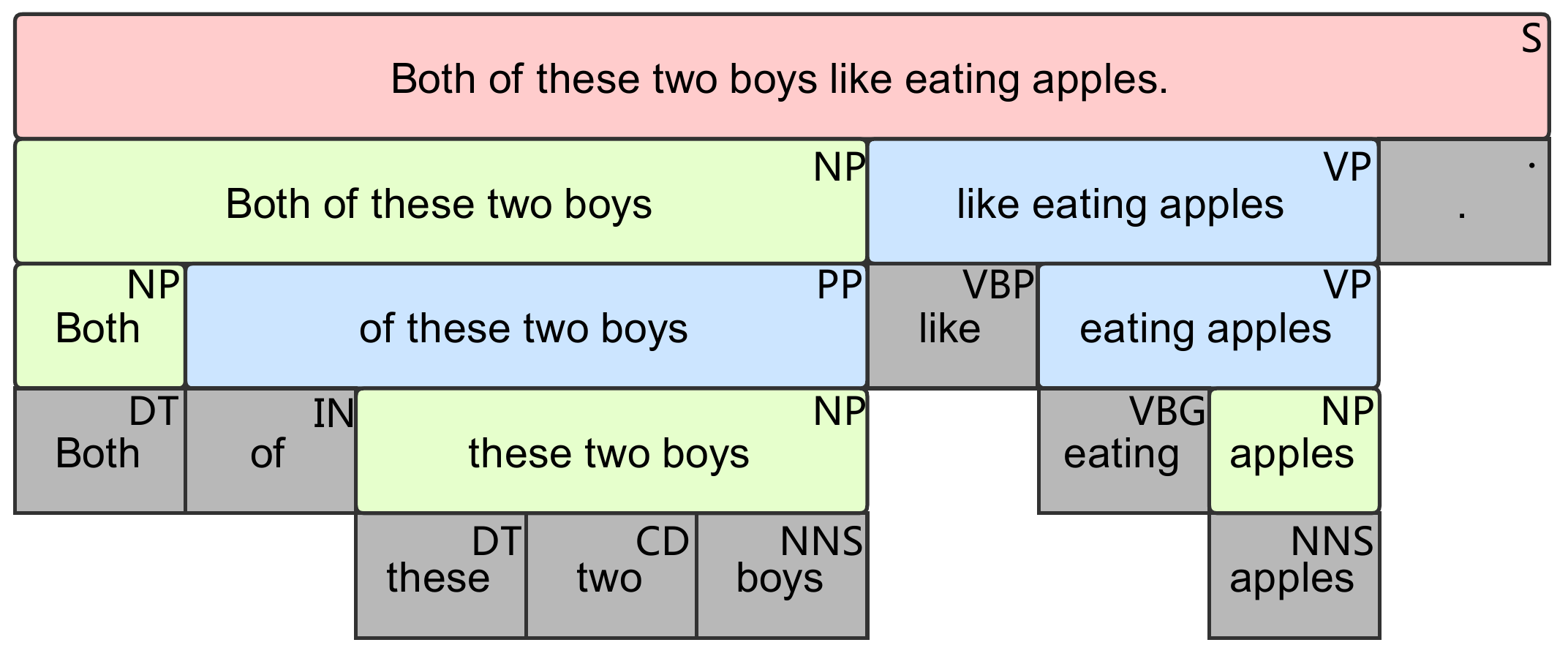}
  \caption{An example of syntactically parsed tree}
  \label{fig:trees}
\end{figure}

In this paper, we will try to exploit the syntactic information, particularly on the linguistic features derived from a syntactic parsed tree for end-to-end TTS. Syntactic parsing is a widely used tool of syntactic analysis. It is also known as "phrase structure parsing", which can describe phrase structure and phrase-level relation between words in a sentence. We have made a systematical study of syntactic parsing from the perspective of TTS, and propose a series of syntactic parsing based features from different viewpoints to help optimize end-to-end TTS model. We use three different test sets to evaluate our models from three aspects, performance on common test set, performance on complex sentence and generalization on pathological test set. Experimental results show that these features are helpful for improving prosody and generalization.

\section{Syntactic Parsing Derived Linguistic Features}
\label{sec:features}

\subsection{Syntactic parsing}
\label{ssec:parsing}

Syntactic parsing decomposes a sentence into its syntactic phrase tree structure. The components in the tree have their corresponding levels and their grammatical roles, e.g. noun phrases and verb phrases. A phrase with more than one word can also be parsed further into sub-phrases until a terminal leaf of a word is reached. Syntactic parsing can be recursively done for an example sentence as shown in Fig.\ref{fig:trees}. In recent years, research on parsing technologies has been greatly advanced and many new parsing algorithms are available , like Probabilistic Context Free Grammar (PCFG) \cite{klein2003accurate}, factored parser \cite{klein2003fast}, Shift-Reduce Parser \cite{zhu2013fast} and Straight to the Tree \cite{shen2018straight}. Improved parsing performance has brought less ambiguous and more stable syntactic analysis. The syntactic parsing model trained with a large text database with rich grammatical structure can provide useful syntactic features to TTS.

In early years, syntactic parsing was mainly used to help building a better rule-based prosody prediction module. For example, \cite{bachenko1986contribution} describes a rule-based system with syntactic parsing to infer prosodic phrasing. Then with the development of statistical parametric speech synthesis, syntactic parsing derived features can be used as a front-end for prosody prediction. In \cite{koehn2000improving},  \cite{yu2013prosodic} and \cite{che2014improving}, features extracted from syntactic parsing for improving prosody prediction are presented. In \cite{zhang2012break}, the authors purpose to build a model which can map from a syntactic tree to a prosodic tree to improve break index labelling.  New studies also tried to use syntactic parsing to improve HMM- or DNN-based acoustic models, e.g.  \cite{yu2013overview}, \cite{dall2016redefining} and \cite{sawada2016nitech}. Experimental results show that syntactic parsing can improve prosody of TTS.

In this section, we will try to make a systematic analysis of syntactic parsing in two aspects, its phrase structure and word relations, and to test deep syntactic parsing derived linguistic features for enhancing TTS performance.

\subsection{Features based on phrase structure}
\label{ssec:structure}

When we investigate the syntactic tree structure in a macroscopic way, the tree describes a phrase structure in multiple levels. The phrase structure controls the syntactic framework of a sentence. The rhythm and intonation of a sentence are intrinsically embedded in the tree-based phrase structure. We adopt the features derived from the phrase structure to characterize the syntactic information:

\begin{itemize}
\setlength{\itemsep}{1pt}
\setlength{\parskip}{1.5pt}
\setlength{\topsep}{0pt}
  \item Part-of-speech (POS) of a word
  \item Phrase labels, e.g. \texttt{S}, \texttt{NP}, \texttt{VP} and \texttt{PP}.
  \item Phrase boundary label for the first word of a phrase
  \item Word's relative position in the phrase. $R_{w} = P_{w} / N_{p}$ \par $P_w$: position of the current word in the current phrase \par $N_{p}$: the number of words in the current phrase
\end{itemize}

For example, The word "like" in Fig.\ref{fig:trees} is the boundary of its parent node which is a \texttt{VP}; its POS is \texttt{VBP}; its relative position in \texttt{S} is $R_w = 5/9$ ("." is a word too); it belongs to higher-level phrases, \texttt{S} and \texttt{VP}.

These features can capture the structure information of a sentence, then affect every word with that information. When we use phrase structure-based features, we need to fix the number of layers and the way to select the specific layers. Different size and methods will have different effect on prosody. In \ref{sssec:exp_phrase_selection}, we will talk about the selection of layers.

\subsection{Features based on word relation}
\label{ssec:relation}

High redundancy in a phrase structure makes it harder to extract useful information for prosody prediction with only limited text data. For this reason, we suggest to refine the features. We all know that the features in ToBI are based on word or lower-level phoneme, like stress, emphasis and break. Therefore, we want to focus on the relation between words in the tree by extracting a few features which can describe both word’s syntactic attributes and the relation between two adjacent words to help learn these features of prosody. We define the features and interpret them with Fig. \ref{fig:distance} as:

\begin{figure}[htp]
  \centering
  \includegraphics[width=8.5cm]{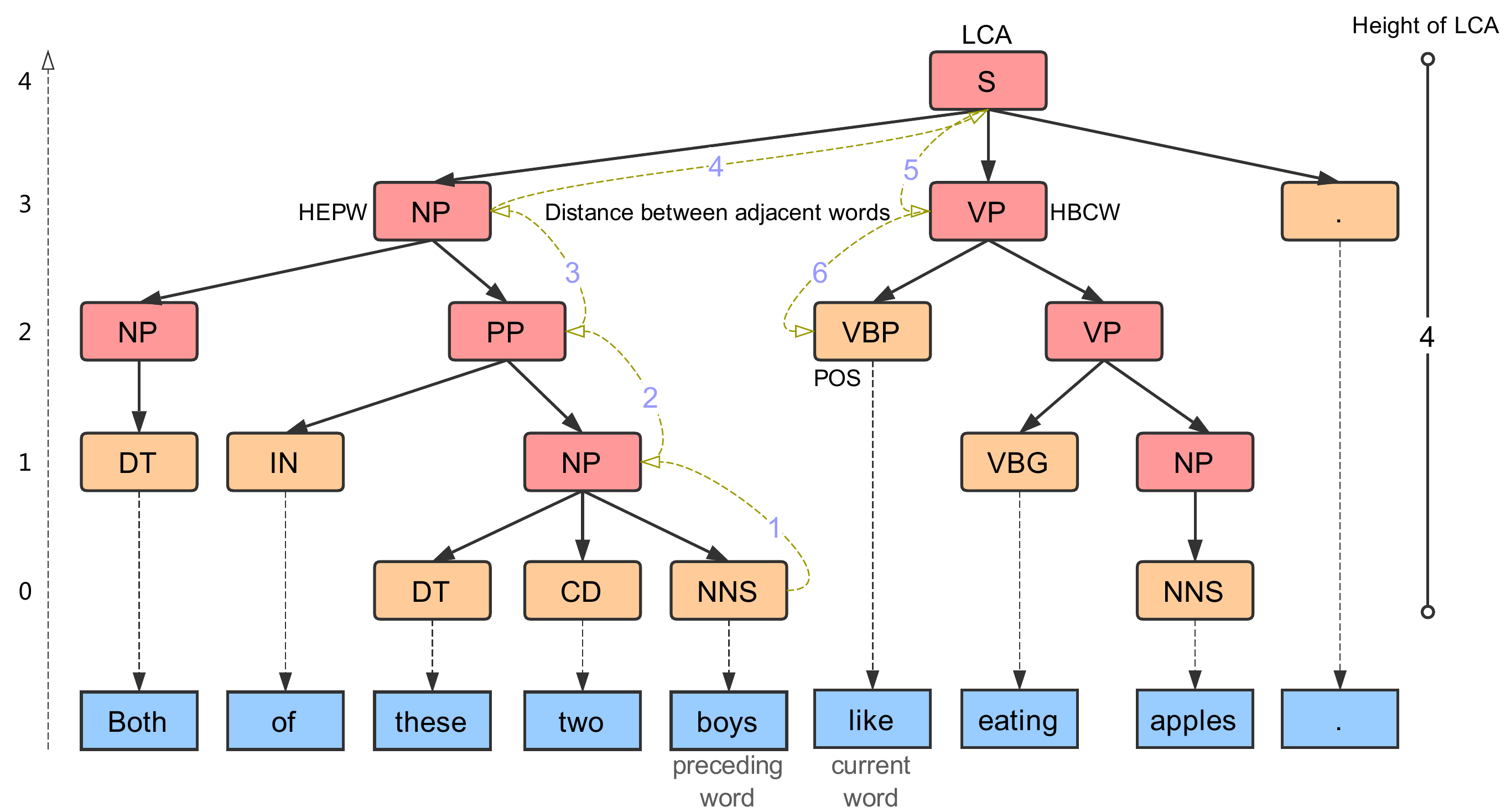}
  \caption{Features based on word relation}
  \label{fig:distance}
\end{figure}

\begin{itemize}[leftmargin=*]
\setlength{\itemsep}{1pt}
\setlength{\parskip}{1.5pt}
  \item Part-of-speech of a word. (Orange part)
  \item Phrases related to the junction of two adjacent words. Breaks or pauses only occur in between two adjacent words. We extend it in this paper, which breaks occur between two adjacent phrases if we set \texttt{NONE} as every word’s first phrase. For example, the phrases at the junction of "boys" and "like" are \texttt{NP} and \texttt{VP}, and the phrases at the junction of "like" and "eating" are \texttt{NONE} and \texttt{VP}. To find these two phrases, we adopt these two criteria:
    \begin{itemize}[leftmargin=*]
      \setlength{\itemsep}{0pt}
      \setlength{\parskip}{1.5pt}
      \item Highest-level phrase beginning with the current word (HBCW).
      \item Highest-level phrase ending with the preceding word (HEPW).
    \end{itemize}
  \item Lowest common ancestor (LCA), the lowest-level node which comprises two adjacent words. In Fig.\ref{fig:distance}, \texttt{S} is the LCA of "boys" and "like", and the 2nd-level \texttt{VP} is the LCA of "eating" and "apples". 
  \item Syntactic distance. It shows the distance between two adjacent words in the tree. Longer distance may lead to higher probability to break with a longer pause. We define the following features related to syntactic distance:
  \begin{itemize}[leftmargin=*]
    \setlength{\itemsep}{0pt}
    \setlength{\parskip}{1.5pt}
    \item Height ($H$): The level in the tree. $H_{l}$, $H_{c}$, $H_{p}$ refer to the level of LCA, current word's POS and preceding word' POS, respectively.
    \item Distance ($D$): The length of shortest path between nodes in the tree (excluding words). $D_{cl}$, $D_{pl}$, $D_{cp}$ refer to the distance between LCA and current POS, LCA and the preceding POS, current POS and preceding POS.
  \end{itemize}
  The $D_{cp}$ of "like" is the length of the shortest path from \texttt{NNS} to \texttt{VBP}. We can add $D_{cl}$ and $D_{pl}$ to get its value. \par($D_{cl} = H_{l} - H_{c}; D_{pl} = H_{l} - H_{p}; D_{cp} = D_{cl} + D_{pl}$)
\end{itemize}

\begin{figure}[htp]
  \centerline{\includegraphics[width=8cm]{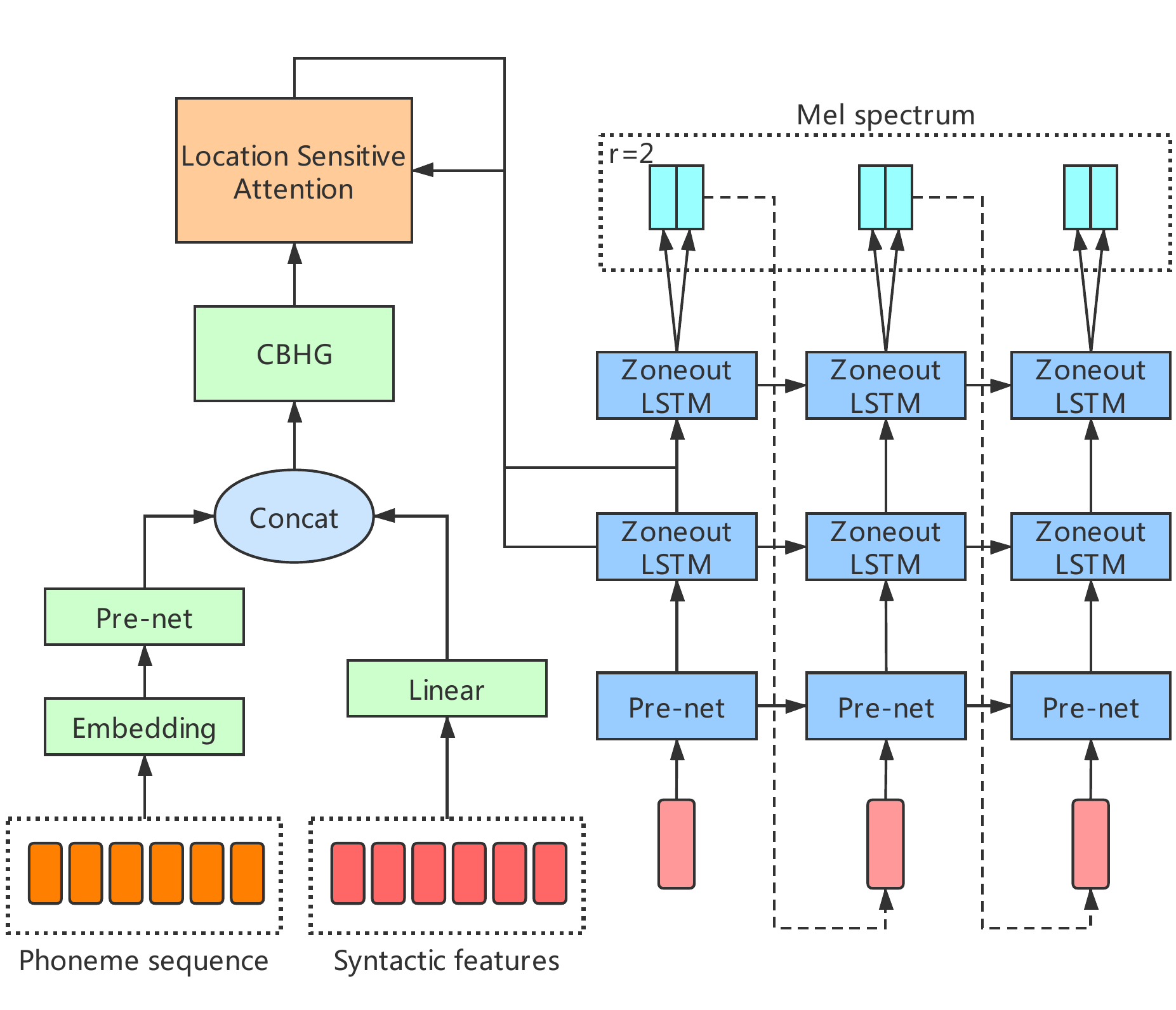}}
  \caption{Model overview}
  \label{fig:res}
  \vspace{-0.3cm}
\end{figure}

\section{Experiments and Results}
\label{sec:experiments}

\subsection{Model architecture \& Training setup}
\label{ssec:setup}

Fig.\ref{fig:res} shows the model architecture. Our model is based on Tacotron1\cite{wang2017tacotron} which can predict Mel spectrum directly from phoneme sequence. We use location sensitive attention which has yielded better result in \cite{shen2018natural}, and replace GRU cells with Zoneout-LSTM\cite{wang2018style} in the decoder to improve the regularization performance. The model output is an $80$-channel, log-mel spectrum, two frames at a time. Finally, we use Giffin-Lim\cite{griffin1984signal} algorithm to synthesize the speech waveform. We up-sample our word-level features from syntactic parsing to phoneme level, and embed them into $256$-dim vectors using a fully-connected layer with ReLU activation, then put them and the outputs of Pre-net together.

We train the end-to-end TTS system with a high-quality American English speech database used in 2011 Blizzard Challenge, which has $16$ hours of speech recorded by a single female speaker. We train these models for 200,000 iterations with a batch size of $128$ distributed across $4$ GPUs with synchronous updates, using $L1$ loss and Adam optimizer with  $\beta_1=0.9$, $\beta_2=0.999$ and a learning rate of $10^{-3}$ exponentially decayed to $10^{-5}$ after $50,000$ iterations. In this study, we use factored parser \cite{klein2003fast} of the Stanford Parser \cite{stanford2011stanford} to extract syntactic trees.

\subsection{Selected features}
\label{ssec:exp_phrase}

\subsubsection{Selected features based on phrase structure}
\label{sssec:exp_phrase_selection}

There are $39$ different POS and $27$ phrase labels used in this study. Hence, a $39 + 29 * N$-dim vector ($27$-dim label + $1$-dim boundary + $1$-dim position for $N$ levels) is used to represent the syntactic information of a word. Because the max depth in our training set is 15, we choose to compare the performance of models using 3, 5, 10, 15 layers in two different, top-down and bottom-up, ways to determine the dimension. We use a test set with 50 sentences to make Comparative Mean Opinion Score (CMOS) test to evaluate the performance on prosody and naturalness of these models. Each pair of samples is judged by 10 native English speakers with a score from -3 to 3.

Table \ref{tab:different_layers} shows the results of subjective preference test. The comparisons show most models achieve similar performance. But their good effect are performed on different sentences. For example, the comparison of $T_5$ and $T_{10}$ show a small CMOS score of $0.03$. But the preferences on these two models are high and similar. It shows that the two features have their own unique effects on different sentences. So we infer that selection of different layers may lead to different performance on the same sentence. Finally we sort these models by CMOS scores and preference, then adopt $T_{10}$ as phrase structure based features (PSF).

\begin{table}[htp]
\centering
\setlength{\belowcaptionskip}{10pt}
\setlength{\tabcolsep}{3mm}
\renewcommand\arraystretch{1.3}
\begin{tabular}{|c|c|c|c|c|c|}
\hline
\multicolumn{5}{|c|}{Preference (\%)} & \multirow{2}{*}{CMOS} \\ \cline{1-5}
\multicolumn{2}{|c|}{Feature A} & Neutral & \multicolumn{2}{c|}{Feature B} &  \\ \hline
$B_5$ & 34.8 & 23.8 & 41.4 & $B_3$ & 0.11 \\ \hline
$B_3$ & 19 & 58.8 & 22.2 & $T_{10}$ & 0.05 \\ \hline
$T_3$ & 35.2 & 23.2 & 41.6 & $T_5$ & 0.10 \\ \hline
$T_{15}$ & 40.4 & 13.2 & 46.4 & $T_5$ & 0.07 \\ \hline
$T_{5}$ & 42.4 & 18.0 & 39.6 & $T_{10}$ & 0.03 \\ \hline
\end{tabular}
\caption{Subjective evaluations on different layers and orders. $T$: top-down, $B$: bottom-up, e.g. $T_5$ refers to top-down 5 levels}
\label{tab:different_layers}
\vspace{-0.3cm}
\end{table}

\subsubsection{Selected features based upon word relation}
\label{sssec:exp_word_relation}

Since all features presented in \ref{ssec:relation} are relevant to prosody generation, we use all of them to represent the word relation. We use one-hot to label POS, HBCW, HEPW and LCA, and adopt $H_l$, $D_{cl}$, $D_{pl}$ and $D_{cp}$ without normalization to represent the syntactic distance. Finally, we set this $124$-dim features as word relation based features (WRF). The dimension is obviously less than phrase structure based features (PSF).

We compare the baseline system (BASE, where only phoneme sequence information is used) with two other systems trained with different features, PSF and WRF. We use three, common, complex and pathological test sets\footnote{Samples are available at \url{https://hhguo.github.io/demo/publications/SyntacticParsing/index.html}} to evaluate the prosody, naturalness, pronunciation clarity and generalization capability of the models in two subjective tests, i.e., preference test and diagnostic intelligibility/naturalness test without using semantically unpredictable sentences (SUS).

\subsection{Comparison}
\label{ssec:comparison}

\subsubsection{Preference test}

The common test set has 50 sentences which are the sentences typical used in news and general conversations. The complex test set consists of 20 sentences which have more complex grammar or longer sentence length. We use these two test sets to make CMOS test to evaluate the performance on prosody and naturalness of the three models. Each case is judged by 20 native English speakers with a score from -3 to 3.

Experimental results in Table \ref{tab:subjective_three_features} show that both features can improve the baseline performance. In these two test sets, syntactic feature based models have larger improvement on the complex test set. These results show that providing syntactic information can significantly help synthesize better speech, especially for sentences with more complex grammar and longer length. Our analysis of these test cases find that the improvement of the effect is mainly on the prosody.

\begin{figure}[htp]
  \centerline{\includegraphics[width=10cm]{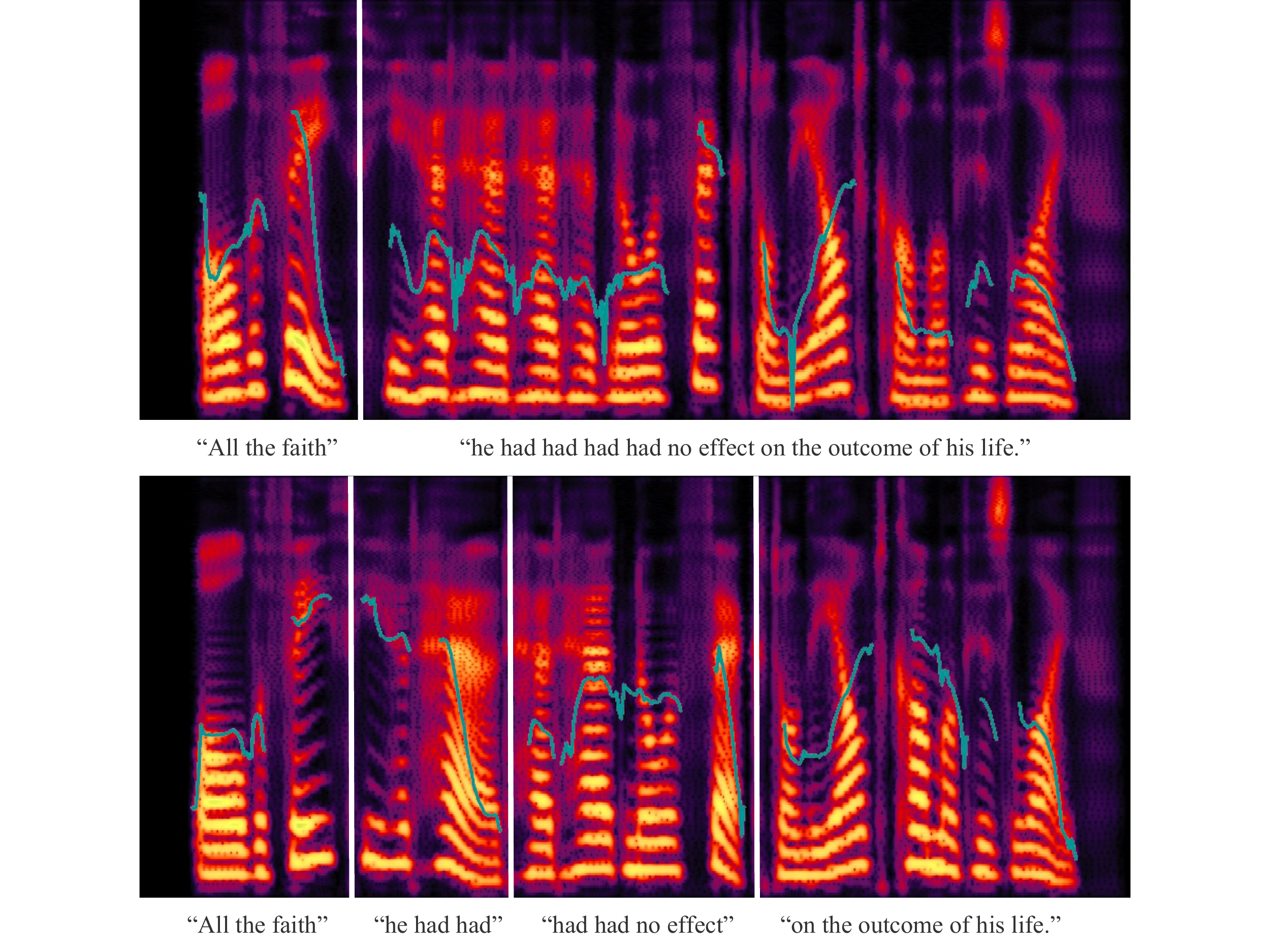}}
  \caption{Comparison of Mel spectrograms synthesized by the baseline model (upper part) and the WRF-based model (lower part). The text below the figure corresponds to the transcript. The white line indicates a pause. The green curve represents the F0 trajectory.}
  \label{fig:sample}
  \vspace{-0.2cm}
\end{figure}

As shown in Fig.\ref{fig:sample}, two samples are synthesized of the same sentence. They have similar performance in spectral clarity, but have obvious difference in prosody. Compared with the sample generated by the baseline model (the upper part), the sample generated by the WRF-based model has better prosody. The syntactic features have helped the model to distinguish the four "had", and insert an appropriate pause after the second "had". Compared with the baseline sample, a better prosody is produced.

Both two features have improve the end-to-end TTS, but WRF yields a higher CMOS score than PSF. It shows that WRF are more effective than PSF. PSF contains more redundant information, which may make model learning less effective.

\begin{table}[htb]
\small
\centering
\setlength{\belowcaptionskip}{10pt}
\setlength{\tabcolsep}{3mm}
\renewcommand\arraystretch{1.3}
\begin{tabular}{|c|c|c|c|c|}
\hline
\multicolumn{4}{|c|}{Preference (\%)} & \multirow{2}{*}{CMOS} \\ \cline{1-4}
Baseline & Neutral & PSF & WRF & \\ \hline
\multicolumn{5}{|c|}{Common Test Set} \\ \hline
17.2 & 58.6 & 24.2 & & 0.122 \\ \hline
21.6 & 38.2 & & 40.2 & 0.258 \\ \hline
\multicolumn{5}{|c|}{Complex Test Set} \\ \hline
7.0 & 70.0 & 23.0 & & 0.230\\ \hline
22.5 & 33 & & 44.5 & 0.400 \\ \hline
\end{tabular}
\caption{Subjective preference of models trained with different input features}
\label{tab:subjective_three_features}
\vspace{-0.2cm}
\end{table}

\subsubsection{Diagnostic intelligibility/naturalness test}

We collected $200$ sentences as a pathological test set, which has richer text content, such as long text, URL, sequence of numbers or characters, abbreviation, e.g. 
\begin{itemize}[leftmargin=*]
\setlength{\itemsep}{1pt}
\setlength{\parskip}{1.5pt}
  \item "You can call me at four two five seven zero three seven three four four or my cell four two five four four four seven four seven four."
  \item "h t t p colon slash slash news dot com dot com slash i slash n e slash f d slash two zero zero three slash f d ."
\end{itemize}
The text and corresponding contexts are not well covered in the training set and the corresponding speech synthesized by the end-to-end TTS are not good. So we use it to evaluate the pronunciation clarity and generalization capability of the models by the corresponding diagnostic intelligibility and naturalness tests. During the test, the listeners are asked to judge if there is any word in the sentence unintelligible or unnatural, then mark them if there are.

Table \ref{tab:intelligibility} shows that the corresponding intelligibility and naturalness rates (\%) of the three models. Around $10\%$ of results synthesized by baseline model have intelligibility or naturalness issues. When the models are trained with the syntactic features, the troublesome issues are significantly improved. It shows that syntactic information is helpful to improve the pronunciation clarity and generalization capability. WRF still yields the best performance, in comparing with PSF or the baseline system as shown in the Table.

\begin{table}[htb]
\small
\centering
\setlength{\belowcaptionskip}{10pt}
\setlength{\tabcolsep}{2.5mm}
\renewcommand\arraystretch{1.3}
\begin{tabular}{|c|c|c|c|c|}
\hline
\multicolumn{4}{|c|}{Diagnostic Intelligibility/Naturalness Rate (\%)} \\ \hline
Case Level & Baseline & With PSF & With WRF  \\ \hline
Intelligible & 92.0 & 96.5 & \textbf{99.0} \\ \hline
Natural & 89.5 & 94.0 & \textbf{97.0} \\ \hline
\end{tabular}
\caption{Diagnostic intelligibility and naturalness test without using SUS sentences on the pathological test set}
\label{tab:intelligibility}
\end{table}

\section{Conclusions}
\label{sec:conclusion}

In this study we investigate syntactic parsing derived features embedded in a parsed tree for improving end-to-end TTS synthesis performance. Two specific features, phrase structure and word relation, are favorably selected to test their effects on prosody prediction, pronunciation clarity, naturalness and generalization of the end-to-end TTS synthesis. Experimental results show that syntactic features can indeed improve the quality of the synthesized speech in its prosody, intelligibility and generalization. The word relation based features (WRF) yield the best performance on three test sets examined.

\vfill\pagebreak

\small
\bibliographystyle{IEEEtran}
\bibliography{refs}

\end{document}